\documentclass[11pt,a4paper]{article}
\usepackage[hyperref]{acl2021}
\usepackage{times}
\usepackage{latexsym}
\usepackage{graphicx}
\usepackage{algorithm}
\usepackage{algorithmic}
\usepackage{caption}
\usepackage{multirow}
\usepackage{amssymb}

\usepackage{microtype}

\aclfinalcopy 

\title{Boosting Span-based Joint Entity and Relation Extraction via Sequence Tagging Mechanism}

\author{Bin Ji, Jing Yang, Jie Yu, Shasha Li, Jun Ma, Huijun Liu \\
  College of Computer, National University of Defense Technology \\
  \texttt{\{jibin,yangjing,yj,shashali,majun,liuhuijun\}@nudt.edu.cn} \\}

\date{}

\begin{document}
\maketitle
\begin{abstract}

Span-based joint extraction simultaneously conducts named entity recognition (NER) and relation extraction (RE) in text span form. Recent studies have shown that token labels can convey crucial task-specific information and enrich token semantics. However, as far as we know, due to completely abstain from sequence tagging mechanism, all prior span-based work fails to use token label information. To solve this problem, we propose Sequence Tagging enhanced Span-based Network (STSN), a span-based joint extraction network that is enhanced by token BIO label information derived from sequence tagging based NER. By stacking multiple attention layers in depth,
we design a deep neural architecture to build STSN, and each attention layer consists of three basic attention units.
The deep neural architecture first learns semantic representations for token labels and span-based joint extraction, and then constructs information interactions between them,  
which also realizes bidirectional information interactions between span-based NER and RE. 
Furthermore, we extend the BIO tagging scheme to make STSN can extract overlapping entity.
Experiments on three benchmark datasets show that our model consistently outperforms previous optimal models by a large margin, creating new state-of-the-art results\footnote{Training code and models will be available at xxx.}.
\end{abstract}

\section{Introduction}

Joint entity and relation extraction aims to simultaneously detect entities and semantic relations among these entities from unstructured texts. 
It serves as a step stone for many downstream tasks such as question answering, knowledge base population, and has recently become a major focus of NLP researches. 

Generally, joint entity and relation extraction can be classified into two categories: sequence tagging based mode \cite{li_ji,mi_ba,ka_ca,wei,lin,zhao} and span-based mode. 
Recently, extensive span-based approaches have been investigated due to its good performance \cite{luan18,di_al,eb_ul,ji}.
Typically, span-based approaches first split text into text spans as candidate entities; then form span pairs as candidate relation tuples; finally, jointly classify spans and span pairs. For example, in Figure 1, \textit{``Jack"}, \textit{``Jack taught"}…are text spans; $<$\textit{``Jack"}, \textit{``Jack taught"}$>$, $<$\textit{``Jack"}, \textit{``Harvard University"}$>$…are span pairs; and \textit{``Jack"} is classified into \texttt{PER}, $<$\textit{``Jack", ``Harvard University"}$>$ is classified into \texttt{Work-for}.
Although these approaches have achieved promising results, they still suffer from insufficient semantics of text spans, caused by excessively relying on the encoding ability of pre-trained language models, e.g., ELMo \cite{elmo}, BERT \cite{bert}.

\begin{figure}[t]
\centering
\includegraphics[width=0.49\textwidth]{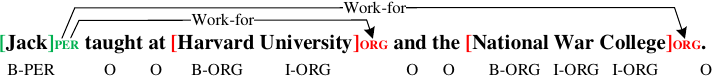} 
\caption{A span-based joint extraction example from CoNLL04, which is tagged by BIO tagging scheme for sequence tagging based NER.}
\label{model2}
\end{figure}

Recent span-based work further enriches semantics by introducing other NLP tasks, e.g., coreference resolution \cite{luan}, event detection \cite{wadden}, but these work needs additional data annotations besides annotations for NER and RE, which do not exist in many datasets.
Actually, prior sequence tagging based work \cite{mi_ba,be,wei,zhao} has shown that token BIO labels for sequence tagging based NER can convey crucial task-specific information and greatly enrich token semantics, which turns out to deliver significant performance improvements. For example, if the model knows that \textit{“Jack”} is tagged with \texttt{PER} label and \textit{“Harvard University”} is tagged with \texttt{ORG} label beforehand, it can easily infer that there may exist a \texttt{Work-for} relation between them. However, to the best of our knowledge, all prior span-based work fails to use token label information, due to that they completely abstain from sequence tagging mechanism. Besides, by leveraging token label information in RE, prior work \cite{mi_ba,be,wei,zhao} constructs an information flow from NER to RE, but fails to construct an information flow from RE to NER.

To address above issues, we propose Sequence Tagging enhanced Span-based Network (STSN), a span-based joint extraction network that can leverage token BIO label information derived from sequence tagging based NER. To build STSN, we design a deep neural architecture by stacking multiple attention layers in depth. 
The deep neural architecture first learns semantic representations of 
token BIO labels (\textbf{``label semantic representations"} for short) for sequence tagging based NER, and token semantic representations for span-based NER and RE, 
and then constructs information interactions among the learned semantic representations. Subsequently, STSN decodes these fully interactive semantic representations.  

Each of the deep stacked attention layers consists of three basic attention units: 
(1) Entity\&Relation to Label Attention (\textbf{E\&R-L-A}), which can feed back task information of span-based NER and RE to label semantic representations, 
aiming to make label semantic representations can better capture task-specific information;
(2) Label to Entity Attention (\textbf{L-E-A}), which can inject NER-specific information derived from label semantic representations into token semantic representations for span-based NER; 
(3) Label to Relation Attention (\textbf{L-R-A}), which can inject RE-specific information derived from label semantic representations into token semantic representations for span-based RE.

It is worth noting that by taking E\&R-L-A as medium, STSN builds bidirectional information interactions between span-based NER and RE,
which will be proven to be of great significance in $\S4.4$.
Besides, L-E-A makes STSN can utilize token label information in span-based NER, which is important but all prior joint extraction work fails to do, as far as we know. 
For example, a model can easily classify \textit{``Harvard University"} to \texttt{ORG} type, if it knows the span is tagged with \texttt{ORG} label beforehand.  
Moreover, to make STSN can extract overlapping entity, we extend the BIO tagging scheme to tag overlapping entity, details will be described in \S{4.1}.
Furthermore,
STSN can leverage fixed-size token label embeddings by concatenation manner.

We conduct extensive experiments on ACE05, CoNLL04 and ADE to evaluate STSN. Experimental results show that STSN overwhelmingly outperforms previous best performed models on above three benchmark datasets, creating new state-of-the-art results.
To the best of our knowledge, STSN is the first span-based joint extraction network that incorporates sequence tagging mechanism, making use of advantages of both span-based and sequence tagging based joint extraction.

\section{Related Works}

\paragraph{Span-based joint extraction.} 
Span-based joint entity and relation extraction has been widely investigated. \cite{luan18} propose the first published span-based model, obtaining span representations the same as \cite{lee}, and sharing them in both NER and RE.
\cite{di_al} realize span-based extraction by obtaining span representations through a BiLSTM over concatenation of ELMo, word and character embeddings. 
\cite{eb_ul} propose SpERT, which takes BERT as backbone and dramatically reduces training complexity by adopting negative sampling. \cite{ji} further improve SpERT by enriching semantic representations with local and global features. Some approaches further promote system performance by incorporating other NLP tasks. \cite{luan} propose DyGIE, a span-based joint extraction model that incorporates coreference resolution. \cite{wadden} introduce event extraction to DyGIE and propose DyGIE++.
Compared to these work, our span-based model incorporates sequence tagging mechanism, and consists of cascaded attention layers to fully leverage token label information, which does not need additional data annotations.

\begin{figure*}[t]
\centering
\includegraphics[width=0.75\textwidth]{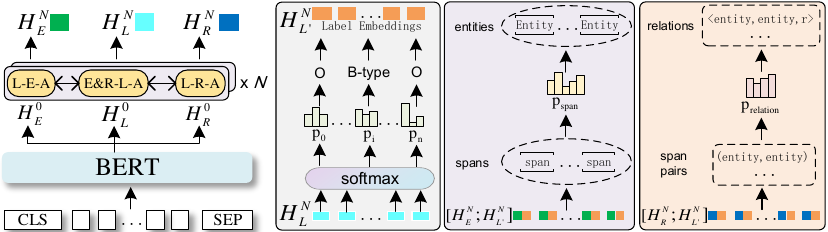} 
\caption{Model architecture of STSN. STSN first generates BERT embeddings for a input sentence and projects them to initial label semantic representations, token representations for span-based NER and RE, i.e., $ H_L^0, H_E^0, H_R^0$; then the deep stacked attention layers fully inject (1) task information derived from $H_E$ and $H_R$ into $H_L$, and (2) task-specific information derived from $H_L$ into $H_E$ and $H_R$, in a recursive way, obtaining the final outputs, i.e., $ H_L^N, H_E^N, H_R^N$; finally, three linear decoders are designed to decode token BIO labels for sequence tagging based NER, span-based NER and RE with $ H_L^N, H_E^N, H_R^N$. Meanwhile, STSN constructs bidirectional information interactions between span-based NER and RE, by taking E\&R-L-A as medium.}
\label{model2}
\end{figure*}

\paragraph{Token label information.} Recent work has proven that token BIO labels for sequence tagging based NER can convey crucial task-specific information for NLP tasks \cite{wa_su,cui}. However, this crucial information has not been fully studied in joint entity and relation extraction. Generally, prior sequence tagging based work \cite{mi_ba,be,wei} explicitly concatenates fixed-size embeddings for token BIO labels to token semantic representations for RE. \cite{zhao} further propose a deep fusion manner.
Limited by the sequence tagging mechanism, these work only leverages token label information in RE, and constructs a unidirectional information flow from NER to RE. 
In contrast, our model is in span-based mode, making it possible to use token label information in both NER and RE. Besides, our model constructs bidirectional information interactions between NER and RE.

\section{Model}

In this section, we will describe the Sequence Tagging enhanced Span-based Network (STSN) in detail, as Figure 2 shows.
For a given sentence $\scriptsize{\mathcal{S} = (t_1,  t_2, t_3,…, t_n)}$, where $t_i$ denotes the $i$-$th$ token in the sentence,
we first generate its semantic representations using pre-trained language models ($\S$3.1); 
then, we construct a deep neural architecture by stacking multiple attention layers in depth and each layer consists of three basic attention units -$ $- Entity\&Relation to Label Attention (E\&R-L-A), Label to Entity Attention (L-E-A) and Label to Relation Attention (L-R-A), which are used to build information interactions between token labels derived from sequence tagging based NER and span-based joint extraction ($\S$3.2); finally, we design three linear decoders for sequence tagging based NER, span-based NER and RE respectively ($\S$3.3).


\subsection{Embedding Layer}


We use BERT \cite{bert} as the default word embedding generator.
For $\mathcal{S}$, BERT first tokenizes it with the WordPiece vocabulary \cite{wordpiece} to get the input sequence. For each sequence element, its input representation is the element-wise addition of WordPiece embedding, positional embedding, and segment embedding. Then, a list of input embeddings $\mathbf{H} \in \mathbb{R}^{len \ast dim}$ are obtained, where \textit{len} is the sequence length and \textit{dim} is the hidden size. A series of pre-trained Transformer blocks \cite{transformer} are then used to project $\mathbf{H}$ into BERT embedding sequence.
$$\scriptsize{\mathbf{E}_\mathcal{S} = \{X_1, X_2, X_3,..., X_{len}\}}$$

Where $X_1$ and $X_{len}$ are the BERT embeddings for the specific [CLS] and [SEP] tokens.

BERT may tokenize a token into several sub-tokens to alleviate out of vocabulary problem. In STSN, we apply $max$-$pooling$ to BERT embeddings for the tokenized sub-tokens of one token to obtain token semantic representation (\textbf{``token representation"} for short), aiming to align token sequence and token representation sequence.  We remove the BERT embeddings for [CLS] and [SEP], and denote the representation sequence for $\mathcal{S}$ as:
$${\hat{\mathbf{E}}_\mathcal{S} = \{\hat{X}_1, \hat{X}_2, \hat{X}_3,..., \hat{X}_{n}\}}$$

Where ${\hat{\mathbf{E}}_\mathcal{S}\in \mathbb{R}^{n*d}}$ and $d$ is the BERT embedding dimension.

\subsection{Attention Layer}

\subsubsection{Deep Stacked Neural Architecture}

We deep stack multiple attention layers to construct a deep neural architecture for STSN, and each of the stacked attention layers consists of three basic attention units, i.e.  E\&R-L-A, L-E-A and L-R-A, as shown in Figure 2.

The deep stacked neural architecture learns three semantic representations, i.e., semantic representations of token BIO labels for sequence tagging based NER (denoted as $H_L$), token representations for span-based NER (denoted as $H_E$) and token representations for span-based RE (denoted as $H_R$), of which the dimensions keep the same. Besides, we denote $H_C$ as the concatenation of $H_E$ and $H_R$, which keeps the same dimension to $H_E$ and $H_R$ via a FFN.
$$H_C = [H_E; H_R]W^C + b^C $$

Where $W^C$, $b^C$ are trainable FFN parameters.

For the first stacked attention layer, $\hat{E_\mathcal{S}}$ is projected into $H_L^0$, $H_E^0$ and $H_R^0$ respectively, which can be formulated as:
$$[Layer]^1 = \left\{ \begin{array}{lrc}
H_C^0 = [H_E^0; H_R^0]W_C^0 + b_C^0 \\
H_L^1 = E\&R$-$L$-$A(H_L^0, H_C^0, H_C^0)\\
H_E^1 = L$-$E$-$A(H_E^0, H_L^1, H_L^1)\\
H_R^1 = L$-$R$-$A(H_R^0, H_L^1, H_L^1) 
\end{array}\right.$$

\begin{figure}[t]
\centering
\includegraphics[width=0.4\textwidth]{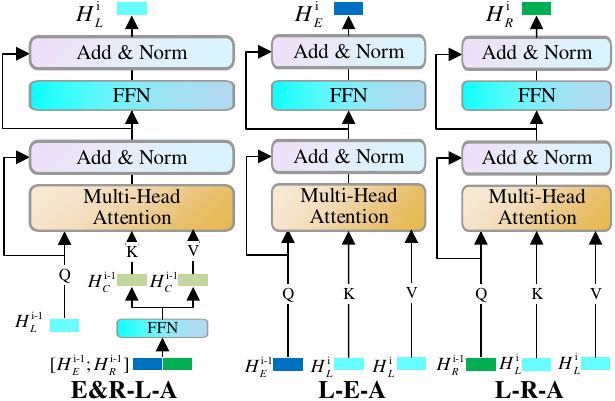} 
\caption{Model architectures of E\&R-L-A, L-E-A and L-R-A. The three units interact with each other through model inputs.}
\label{model2}
\end{figure}

Where $W_C^0 \in \mathbb{R}^{2d*d}$ and $b_C^0 \in \mathbb{R}^d$ are trainable FFN parameters. We then pass the layer outputs, i.e., $ H_L^1$, $ H_E^1$, $ H_R^1$, to the next layer and repeat this procedure in a recursive manner, until we obtain the outputs of the $N$-$th$ layer, i.e., $ H_L^N$, $ H_E^N$ and $ H_R^N$, which are taken as the final semantic representations. Now, $ H_E^N$ and $ H_R^N$ fully contain task-specific information and are suitable for span-based NER and RE. 

As shown in Figure 2,  in each stacked attention layer, we construct bidirectional information interactions among the three basic attention units. Specifically, E\&R-L-A and L-E-A can directly interact with each other, as well E\&R-L-A and L-R-A. Therefore, by taking E\&R-L-A as medium, L-E-A and L-R-A can interact with each other too, which constructs bidirectional information interactions between span-based NER and RE in essence. Besides, L-E-A makes STSN can utilize token label information in span-based NER.

\subsubsection{Three basic Attention Units}

As Figure 3 shows, the three basic attention units share a general neural architecture, but differ in model inputs. We first describe the general architecture, then introduce their implementation details.

\paragraph{General architecture.} The general architecture has two sub-layers: the first is a multi-head attention, the second is a position-wise Feed Forward Network (FFN). A residual connection is adopted around each of the two sub-layers, followed by layer normalization. The general architecture is similar to Transformer encoder layer, but they differ in model inputs.

Multi-head attention has proven to be effective for capturing long-range dependencies by explicitly attending to all positions. Therefore, we apply the multi-head attention to capture task information and task-specific information.  The multi-head attention owns a serious of $h$ parallel heads and requires three inputs, i.e., query, key and value. 
$$head_i = softmax(\frac{(QW_i^Q)(KW_i^K)^T}{\sqrt{d/h}}(VW_i^V))$$
$$I = Concat(head_1,….head_h)W^o$$

Where $\{Q, K, V\} \in \mathbb{R}^{n*d}$ are query, key and value; $\{W_i^Q, W_i^K, W_i^V\} \in {\mathbb{R}^{d*d/h}}$, $W^o \in \mathbb{R}^{d*d}$ are trainable model parameters; and $I \in \mathbb{R}^{n*d}$ is the output.
The multi-head attention learns the pairwise relationship between the query and key, and finally outputs the captured information by weighted summation across all instances.
Then the residual connection conducts element-wise addition of the captured information and query, accomplishing information injection in a shallow manner.

The position-wise FFN contains two linear transformations with a ReLU activation in between. 
$$FFN(I) = max(0, IW_1 + b_1)W_2 + b_2$$

Where $\{W_1, W_2\} \in \mathbb{R}^{d*d}$ and $\{b_1, b_2\} \in \mathbb{R}^d$ are trainable FFN parameters. By conducting linear transformation on the outputs of multi-head attention, FFN projects the captured information to the representation space of $I$, which further accomplishes information injection in a deep manner.

\paragraph{E\&R-L-A.} E\&R-L-A (see Figure 3) takes $H_L$ as query, $H_C$ as key and value respectively, 
and can feed back task information of span-based NER and RE, which is captured from $H_C$, to $H_L$, aiming to make $H_L$ can better learn task-specific information.



\paragraph{L-E-A and L-R-A.} L-E-A (see Figure 3) takes $H_E$ as query, $H_L$ as key and value respectively, and is capable of injecting task-specific information of span-based NER, which is captured from $H_L$, into $H_E$. 
Similarly, L-R-A (see Figure 3) takes $H_R$ as query, $H_L$ as key and value respectively, and can inject RE-specific information into $H_R$.

\subsection{Decoding Layer}
We propose three linear decoders for sequence tagging based NER, span-based NER and RE respectively. 
\paragraph{Decoder for sequence tagging based NER.} This decoder first utilizes a FFN to reduce label semantic representation, i.e., $H_L^N$, to space of token BIO label size; then utilizes $softmax$ to calculate probability distributions on the label size space.
$$\hat{y}^L = softmax(H_L^N W^L + b^L)$$

Where $ W^L \in \mathbb{R}^{d*l}$ and $ b^L \in \mathbb{R}^{l}$ are trainable FFN parameters; $l$ is the token BIO label size.
The training objective is to minimize the cross-entropy loss.
$$\scriptsize{\mathcal{L}_L = -\frac{1}{M_{L}} \sum \limits_{i=1}^{M_{L}}{y}_i^{{L}}\log\hat{y}_i^{{L}}} $$

Where ${y}^{{L}}$ is the one-hot vector of gold token BIO label for sequence tagging based NER; ${M_{L}}$ is the number of token instances. For each token BIO label, we maintain a label embedding with fixed size (denoted as $H^N_{L'}$) for it, which is trained during model training. We use gold token BIO labels during model training and predicted labels during inference.

\paragraph{Decoder for span-based NER.}

We first add \texttt{NoneEntity} type to the pre-defined entity types. Spans will be decoded into \texttt{NoneEntity} space if they are not predicted as entities.
We formulate an arbitrary span from $\mathcal{S}$ to help introduce the rest:
$\scriptsize{\mathbf{s} = (t_i, t_{i+1}, t_{i+2}...,t_{i+j})}$.

We obtain span representation by concatenating token representations of span head and tail, and a span width embedding, following \cite{luan18}.
$$E_\mathbf{s} = [H_{E',i}^N; H_{E',i+j}^N; W_{j+1}]$$

Where $H_{E',i}^N$ and $H_{E',i+j}^N$ are from [$ H_{E}^N; H^N_{L'}$], which is the concatenation of $ H_{E}^N$ and $H^N_{L'}$; $W_{j+1}$ is a span width embedding with fixed size, which is trained during model training.

$ E_\mathbf{s} $ first passes through a FFN, then is fed into ${softmax}$, which yields a posterior on the space of pre-defined entity type size.
$$\scriptsize{\hat{y}^{\mathbf{s}} = softmax({E}_\mathbf{s}W^\mathbf{s} + b^\mathbf{s})}$$

Where $W^\mathbf{s}$ and $b^\mathbf{s}$ are trainable FFN parameters. 
The training objective is to minimize the cross-entropy loss.
$$\scriptsize{\mathcal{L}_E = -\frac{1}{M_{E}} \sum \limits_{i=1}^{M_{E}}{y}_i^{\mathbf{s}}\log\hat{y}_i^{\mathbf{s}}} $$

Where ${y}^{\mathbf{s}}$ is the one-hot vector of gold span type; ${M_{E}}$ is the number of span instances.

\paragraph{Decoder for span-based RE.} We first formulate an arbitrary span pair from $\mathcal{S}$: 
$\mathbf{r} = <\textbf{s}_1, \textbf{s}_2>$,
where $\mathbf{s}_1$ and $\mathbf{s}_2$ are two text spans. Then we obtain relation representation by concatenation manner. 
$$E_\mathbf{r} = [E_{\mathbf{s}_1}; E_{\mathbf{s}_2}; C_\textbf{r}]$$

Where $E_{\mathbf{s}_1}$ and $E_{\mathbf{s}_2}$ are semantic representations of $\textbf{s}_1$ and $\textbf{s}_2$. It is worth noting that we obtain $E_{\mathbf{s}_1}$ and $E_{\mathbf{s}_2}$ with [$H_{R}^N; H^N_{L’}$] in the same way as in the NER decoder.
Following \cite{eb_ul}, $C_\textbf{r}$ is the semantic representation of relation context, which is obtained with [$H_{R}^N; H^N_{L’}$] as well. 

$ E_\mathbf{r} $ first passes through a FFN, then is fed into ${sigmoid}$, which yields probability distributions on the space of pre-defined relation type size.
$$\scriptsize{\hat{y}^{\mathbf{r}} = \sigma({E}_\mathbf{r}W^\mathbf{r} + b^\mathbf{r})}$$

Where $W^\mathbf{r}$ and $b^\mathbf{r}$ are trainable FFN parameters. 
Given a confidence threshold $\alpha$, any relation with a $score \ge \alpha$ is considered activated. If none is activated, the model assumes that the span pair holds no pre-defined relation.
The training objective is to minimize the binary cross-entropy loss.
$$\scriptsize{\mathcal{L}_R = -\frac{1}{M_{R}} \sum \limits_{i=1}^{M_{R}}(y_i^\mathbf{r}\log\hat{y}_i^\mathbf{r} + (1- y_i^\mathbf{r}) \log(1-\hat{y}_i^\mathbf{r}))}$$

Where ${y}^{\mathbf{r}}$ is the one-hot vector of gold relation type for span pair; ${M_{R}}$ is the number of span pair instances.

Finally, we optimize the following joint objective function during model training.
$$\mathcal{L}_{joint}(W;\theta) = \mathcal{L}_L + \mathcal{L}_E + \mathcal{L}_R$$

\begin{figure}[t]
\centering
\includegraphics[width=0.4\textwidth]{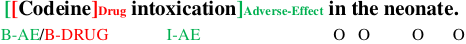} 
\caption{An example of overlapping entities from ADE, tagged by the extended BIO tagging scheme.}
\label{model2}
\end{figure}

\section{Experiments}
\subsection{Experimental Setup}

\paragraph{Datasets.}We evaluate our model on ACE05 \cite{ace}, CoNLL04 \cite{conll} and ADE \cite{ade}.  For ACE05, we use the same entity and relation types, data splits, and pre-processing as \cite{luan,wa_lu}; for CoNLL04, we use the same train-test split as \cite{zhao,wa_lu}; for ADE, we maintain two dataset versions following \cite{eb_ul,wa_lu}, one is the full ADE dataset with 119 instances that contain overlapping entity, the other without these instances. 

\paragraph{Extended BIO tagging scheme.} Aiming to make STSN can extract the overlapping entities from the full ADE dataset, we extend the BIO tagging scheme. We first give two term definitions:  

\begin{itemize}
\item \textbf{two-fold overlapping entities} is a pair of overlapping entities that the overlapping tokens are only contained in the two entities.
\item \textbf{preceding entity} is distinguished by entity head location, while if two entities have the same head location, the entity with longer length is chosen as the preceding entity.
\end{itemize}

As Figure 4 shows, “Codeine” and “Codeine intoxication” are two-fold overlapping entities, and “Codeine intoxication” is the preceding entity.

Detailed tagging rule is that we first tag the preceding entity with BIO tagging scheme; then for the overlapping entity, we append its BIO labels to existing labels, separated by “/”.  
For example, we tag the overlapping “Codeine” with “B-AE/B-DRUG”.  

As all overlapping entities in the full ADE dataset are two-fold, we tag the dataset with the extended BIO tagging scheme.  For other datasets, we tag them with BIO tagging scheme.

\paragraph{Evaluation measures.}We use the F1 measure to evaluate model performance. For NER, a predicted entity is considered correct if both its type and boundaries (span head for ACE05) match ground truth. For RE, we adopt two evaluation measures: a predicted relation is considered correct if its relation type and boundaries of the two entities match ground truth (denoted as RE); while RE+ needs both relation type and the two entities all match ground truth.  More discussion of the evaluation settings can be found in \cite{wa_lu}.

\paragraph{Implementation details.}
For all datasets, we build STSN by stacking 4 attention layers in depth, and evaluate it upon \texttt{bert-base-cased} \cite{bert} and \texttt{albert-xxlarge-v1} \cite{albert} with a single NVIDIA RTX 3090 GPU. We optimize STSN using AdamW for 100 epochs with a learning rate of 5e-5, a linear scheduler with a warmup ratio of 0.1 and a weight decay of 10-2. We set the training batch size to 4, dimensions of $W_{j+1}, H_{L'}^N$ to 150, $h$ for multi-head attention to 8, span width threshold to 10 and relation threshold $\alpha$ to 0.4.
We adopt negative sampling strategy following \cite{eb_ul}.
For ACE05 and CoNLL04, we run STSN for 20 runs and report the averaged F1 of the top-5 runs. For ADE, we adopt 10-fold cross validation, run each fold for 20 runs and report the averaged F1 of the top-5 runs.

\subsection{Main Results}
Table 1 and Table 2 compare performances of STSN with previous optimal results. We report the F1 scores for a fair comparison with prior work. Our BERT-based model significantly outperforms all previous work including models built upon ALBERT, where our model performances are further improved by using ALBERT. For NER, our best model delivers absolute F1 gains of +1.7, +1.5(micro)/+2.7(macro), +1.9 on ACE05, CoNLL04 and ADE(without overlap) respectively, while better ones for RE, by +3.5(RE)/+3.7(RE+), +2.6(micro)/+2.9(macro) and +4.2 respectively.  
All these overwhelming gains demonstrate the effectiveness of leveraging token label information in span-based NER and RE, as well as the bidirectional information interactions between the two tasks.

\begin{table}
\centering
\begin{tabular}{lllll}
\hline
                     Model                                                & \multicolumn{1}{c}{NER}   & \multicolumn{1}{c}{RE}   & \multicolumn{1}{c}{RE+}   \\ \hline
Li and Ji \shortcite{li_ji} $\vartriangle$   			& 80.8  & 52.1 & 49.5  \\
                           Miwa \textit{et al.} \shortcite{mi_ba}               							& 83.4  & \multicolumn{1}{c}{-}    & 55.6  \\
                           Katiyar \textit{et al.} \shortcite{ka_ca}            							& 82.6  & 55.9 & 53.6  \\
                           Sun \textit{et al.} \shortcite{sun}                      									& 83.6  & \multicolumn{1}{c}{-}    & 59.6  \\
                           Li \textit{et al.} \shortcite{li} \dag                        										& 84.8  & \multicolumn{1}{c}{-}    & 60.2  \\
                           Dixit and Al \shortcite{di_al}  $\perp$             		     							& 86.0  & 62.8 & \multicolumn{1}{c}{-}     \\
                           Luan \textit{et al.} \shortcite{luan}                     								& 88.4  & 63.2 & \multicolumn{1}{c}{-}     \\
                           Wadden \textit{et al.} \shortcite{wadden}  \dag                 							& 88.6  & 63.4 & \multicolumn{1}{c}{-}     \\
							  Wang \textit{et al.} \shortcite{wa_su}  \dag                 							& 87.2  & 66.7 & \multicolumn{1}{c}{-}     \\
                           Lin \textit{et al.} \shortcite{lin}  \dag                      									& 88.8  & 67.5 &       \\
                           Wang and Lu \shortcite{wa_lu}  \ddag                								& 89.5  &{67.6} & 64.3  \\
                           Ji \textit{et al.} \shortcite{ji}   \dag                      										& {89.6}  &\multicolumn{1}{c}{-}     &{65.2}   \\ \cline{1-4} 
                           ours  \dag                                                             						&90.2       & 68.1     & 66.9      \\                     
				    		  ours  \ddag                              	&\textbf{91.3}       & \textbf{71.1}     & \textbf{68.9}      \\ \hline
\end{tabular}
\caption{Main results on ACE05 using micro-averaged F1. $\perp$: LSTM + ELMo; \dag: bert-base-cased (or -uncased); \ddag: albert-xxlarge-v1.}
\end{table}

\begin{table}
\centering
\begin{tabular}{lllll}

\hline
                      & Model               & \multicolumn{1}{c}{NER}   & \multicolumn{1}{c}{RE+}   \\ \hline
\multirow{14}{*}{\rotatebox{90}{CoNLL04}} & Miwa et al. \shortcite{mi_sa} $\vartriangle$              	& 80.7      & 61.0  \\
                          & Bekoulis \textit{et al.} \shortcite{be} $\blacktriangle$															& 83.9      & 62.0  \\                   
                          & Nguyen \textit{et al.}\shortcite{ng_ve} $\blacktriangle$          										& 86.2    & 64.4  \\
                          & Zhang \textit{et al.} \shortcite{zhang} $\vartriangle$                  												& 85.6      & 67.8  \\
                          & Li \textit{et al.} \shortcite{li} $\vartriangle$ \dag                       		 					& 87.8   & 68.9  \\
                          & Eberts \textit{et al.} \shortcite{eb_ul} $\vartriangle$ \dag             												& 88.9     & 71.5  \\
                          & Eberts \textit{et al.} \shortcite{eb_ul} $\blacktriangle$ \dag             											& 86.3     & 72.9  \\
                          & Zhao \textit{et al}. \shortcite{zhao} $\vartriangle$ $\perp$                    												& 90.6     & 73.0 \\
                          & Wang and Lu \shortcite{wa_lu} $\vartriangle$ \ddag                  											& 90.1     & 73.6  \\
                          & Wang and Lu \shortcite{wa_lu} $\blacktriangle$ \ddag                  											& {86.9}      & 75.4  \\
                          & Ji \textit{et al.} \shortcite{ji} $\vartriangle$ \dag                 							& {90.2}     & {74.3} \\ \cline{2-4} 
                          & ours $\vartriangle$  \dag                                                   												& 90.8        &75.2       \\
                          & ours $\vartriangle$ \ddag                                          	& \textbf{91.7}            &\textbf{76.9}       \\ \cline{2-4}
							 & ours $\blacktriangle$ \dag                                                           										& 88.3       &77.1       \\ 
				     		
                          & ours $\blacktriangle$ \ddag                  						& \textbf{89.6}            &\textbf{78.3}       \\ \hline
\multirow{10}{*}{\rotatebox{90}{ADE}}     & Li \textit{et al.} \shortcite{li16} $\blacktriangle$                      	& 79.5      & 63.4  \\
                          & Li \textit{et al.} \shortcite{li17} $\blacktriangle$                      										& 84.6       & 71.4  \\
                          & Bekoulis \textit{et al.} \shortcite{be} $\blacktriangle$                  										& 86.4       & 74.6  \\                          
                          & Eberts \textit{et al.} \shortcite{eb_ul} $\blacktriangle$ \dag            									& 89.3       & 79.2  \\
				    & Eberts \textit{et al.} \shortcite{eb_ul} $\blacktriangle\spadesuit$ \dag            									& 89.3        & 78.8  \\
                          & Wang and Lu \shortcite{wa_lu} $\blacktriangle$ \ddag                  									& 89.7        & 80.1  \\
                          & Zhao \textit{et al.} \shortcite{zhao} $\blacktriangle$ $\perp$                    										& 89.4      & 81.1  \\
                          & Ji \textit{et al.} \shortcite{ji} $\blacktriangle$ \dag                         											& 90.6       & 80.7 \\ \cline{2-4} 
                          & ours $\blacktriangle$ \dag                                                             								&91.4        &83.2       \\                         
				    & ours $\blacktriangle$ \ddag                                               	&\textbf{92.5}           &\textbf{84.9}       \\  \cline{2-4} 
					& ours $\blacktriangle\spadesuit$ \dag                                                             								&91.0       &83.4       \\                         
				    & ours $\blacktriangle\spadesuit$ \ddag                                               	&\textbf{91.9}        &\textbf{84.5}       \\ \hline

\end{tabular}
\caption{Main results on CoNLL04 and ADE. $\vartriangle$: micro-averaged F1; $\blacktriangle$: macro-averaged F1; $\spadesuit$: with  overlap; $\perp$: LSTM + ELMo; \dag: bert-base-cased (or -uncased); \ddag: albert-xxlarge-v1.}
\end{table}

We evaluate STSN for extracting overlapping entity on the full ADE dataset. Experimental results are shown in Table 2, marked by $\spadesuit$. Compared to \cite{eb_ul}, our model delivers overwhelming performance gains. Specifically, on NER and RE, our BERT-based model delivers absolute F1 gains of +1.7 and +4.6, while our ALBERT-based model delivers +2.6 and +5.7. These gains validate the effectiveness of both the proposed extended BIO tagging scheme and STSN.

\subsection{Analysis}

Following established line of work \cite{eb_ul,zhao}, we inspect detailed model performance in two aspects by taking SpERT \cite{eb_ul} as baseline. For fair comparison, we utilize the BERT-based STSN.

\begin{figure}[t]
\centering
\includegraphics[width=0.8\columnwidth]{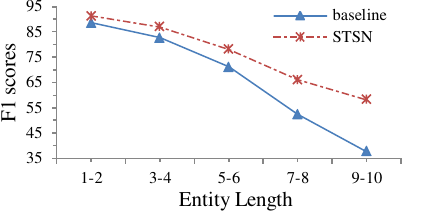}
\caption{NER performance comparisons of the baseline and STSN under different grouped entity lengths on the ACE05 dev set.}
\label{fig1}
\end{figure}

\begin{figure}[t]
\centering
\includegraphics[width=0.98\columnwidth]{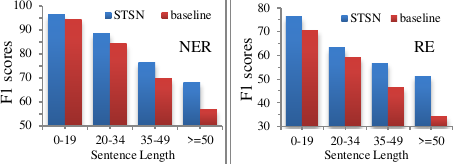}
\caption{NER and RE performances of the baseline and STSN under different grouped sentence lengths on the ACE05 dev set.}
\label{fig1}
\end{figure}

\subsubsection{Performance against Entity Length}
Figure 5 shows NER performance comparisons of the baseline and STSN under different entity distances on the ACE05 dev set. We partition entity lengths, which is restricted by span width threshold, into 1-2, 3-4, 5-6, 7-8, 9-10. We can observe that STSN consistently outperforms the baseline under all length intervals. Moreover, absolute F1 gains achieved by STSN are further enhanced when entity length increases. In particular, STSN obtains 13.7 and 20.5 absolute F1 gains when entity length is 7-8 and 9-10 respectively, demonstrating that STSN is more effective in terms of long entities.

\subsubsection{Performance against Sentence Length}
We analyze performances of the baseline and STSN under different grouped sentence lengths on the ACE05 dev set. As Figure 6 shows, we partition the sentence length into 0-19, 20-34, 35-49, $\geq$50. We can observe that STSN performs way better than the baseline under all sentence lengths. Moreover, the gains achieved by STSN are further enhanced when the sentence length increases. In particular, STSN outperforms the baseline by 11.1 and 16.8 absolute F1 scores on NER and RE, when the sentence length is greater than or equal to 50. These results demonstrate that STSN is more effective in terms of long sentences. 

\begin{table}[]
\centering
\begin{tabular}{lcc}
\hline
Method      & \begin{tabular}[c]{@{}c@{}}Entity\\ (F1)\end{tabular} & \begin{tabular}[c]{@{}c@{}}Relation\\ (F1)\end{tabular} \\ \hline
STSN +   &                                                       &                                                         \\
\quad \texttt{1 AttentionLayer}     	  &87.8                                                      &59.3                                                          \\
\quad \texttt{2 AttentionLayers}  	  &88.6                                                       &59.7                                                        \\
\quad \texttt{3 AttentionLayers} 	  &89.3                                                       &62.1                                                        \\
\quad \texttt{4 AttentionLayers} 	  &\textbf{89.2}                                                       &\textbf{62.8}                                                         \\
\quad \texttt{5 AttentionLayers} 	  &89.0                                                       &62.6                                                         \\
\quad \texttt{6 AttentionLayers}        &89.2                                                       &61.9                                                         \\ 
 \hline
\end{tabular}
\caption{Ablations of attention layers on ACE05 dev set.}
\end{table}

\begin{table}[]
\centering
\begin{tabular}{lcc}
\hline
Method      & \begin{tabular}[c]{@{}c@{}}Entity\\ (F1)\end{tabular} & \begin{tabular}[c]{@{}c@{}}Relation\\ (F1)\end{tabular} \\ \hline
STSN   &89.2                                                       &62.8                                                         \\
\quad \texttt{-Label\_Info}        &87.2                                        &58.6   \\                                                   
\quad \texttt{-Bi\_Interact} &88.4                                                       &60.6 \\
 \hline
\end{tabular}
\caption{Ablations of model components on ACE05 dev set.}
\end{table}

\subsection{Ablation Study}
We conduct ablation studies on the BERT-based STSN, and use the ACE05 dev set to report ablation results.

\paragraph{Ablations of attention layers.} We conduct ablations of attention layers by deep stacking different amounts of attention layers in STSN. Table 3 shows ablation results, from which we can conclude that: (1) the model with four stacked layers performs the best; (2) the model consists of one attention layer performs the worst, due to that one layer cannot fully inject token label information into semantic representations for span-based joint extraction; (3) as the model depth increases, model performance first significantly increases and then slightly decreases. This is because that deep models make it easier to fully inject information, while much deeper models tend to inject more noise, which imposes negative impacts on model performance.

\paragraph{Ablations of model components.}
Table 4 reports ablation results, where 
(1) \texttt{-LabelEmbedding} denotes ablating token label embeddings, which is realized by removing $H^N_{L'}$ from decoders for span-based NER and RE; 
(2) \texttt{-E\&R-L-A} denotes ablating the bidirectional information interactions between span-based NER and RE, which is realized by making E\&R-L-A take $H_L$ as query, key and value respectively in all stacked attention layers;
(3) \texttt{-AttentionLayer} denotes ablating all stacked attention layers, where ${\hat{\mathbf{E}}_\mathcal{S}}$ is directly projected into $H_L^N$, $H_E^N$ and $H_R^N$. 

Ablation results show that: (1) token label embeddings impose minor positive impacts on model performance, by delivering +0.4 and +0.5 absolute F1 gains on NER and RE respectively; (2) bidirectional information interactions significantly improve model performance, by delivering +0.8 and +2.2 absolute F1 gains on NER and RE respectively; (3) stacked attention layers boost model performance, delivering +2.0 and +4.2 absolute F1 gains on NER and RE respectively.

Reasons can be summarized as that for span-based NER and RE:  
(1) the deep stacked attention layers can fully inject token label information into semantic representations for them, which greatly enriches their semantics. While compared to above deep fusion way, shallow concatenation way achieved by concatenating token label embeddings plays a negligible role in enriching their semantics; 
(2) the bidirectional information interactions between the two tasks can greatly enrich their semantics, by fully leveraging task-specific information of each other.


\section{Conclusion}
In this paper, we propose the Sequence Tagging enhanced Span-based Network (STSN) for joint entity and relation extraction. By adopting a deep stacked neural architecture composed of three basic attention units, STSN can leverage token BIO label information in span-based joint entity and relation extraction. Moreover, the deep neural architecture realizes bidirectional information interactions between span-based NER and RE, which greatly promotes information interactions. 
Furthermore, we extend the BIO tagging scheme to make STSN can extract overlapping entity. 
Experiments on three benchmark datasets show that our model overwhelmingly outperforms other competing approaches, creating new state-of-the-art results.

\bibliographystyle{acl_natbib}
\bibliography{anthology,acl2021}


\end{document}